\documentclass[a4paper,conference]{IEEEtran}

\usepackage{graphicx}
\usepackage{comment}
\usepackage{multicol}
\usepackage{float}
\usepackage{lipsum}
\usepackage{algorithm}
\usepackage{algpseudocode}
\usepackage{tabularx}
\usepackage{amsmath}
\usepackage{xcolor}
\usepackage{hyperref} 
\hypersetup{pdfborder={0 0 1},
    colorlinks=true,                
    breaklinks=true,                
    urlcolor= black, 
    anchorcolor = blue,
    linkcolor= red,                 
    bookmarksopen=false,
    filecolor=black,
    citecolor=green,
    linkbordercolor=green,
    citebordercolor=green,
}

\ifCLASSINFOpdf
\else
\fi

\begin{document}

\title{Dynamic Rectification Knowledge Distillation}

\author{\IEEEauthorblockN{Fahad Rahman Amik, Ahnaf Ismat Tasin, Silvia Ahmed, M. M. Lutfe Elahi, Nabeel Mohammed}
\IEEEauthorblockA{\textit{Dept of  Electrical and Computer Engineering} \\
\textit{North South University, Bashundhara}\\
Dhaka 1229, Bangladesh \\
\{fahad.rahman1, ahnaf.ismat, silvia.ahmed, lutfe.elahi, nabeel.mohammed\}@northsouth.edu}}

\maketitle

\begin{abstract}
Knowledge Distillation is a technique which aims to utilize dark knowledge to compress and transfer information from a vast, well-trained neural network (teacher model) to a smaller, less capable neural network (student model) with improved inference efficiency. This approach of distilling knowledge has gained popularity as a result of the prohibitively complicated nature of such cumbersome models for deployment on edge computing devices. Generally, the teacher models used to teach smaller student models are cumbersome in nature and expensive to train. To eliminate the necessity for a cumbersome teacher model completely, we propose a simple yet effective knowledge distillation framework that we termed \emph{Dynamic Rectification Knowledge Distillation} (DR-KD). Our method transforms the student into its own teacher, and if the self-teacher makes wrong predictions while distilling information, the error is rectified prior to the knowledge being distilled. Specifically, the teacher targets are dynamically tweaked by the agency of ground-truth while distilling the knowledge gained from traditional training. Our proposed DR-KD performs remarkably well in the absence of a sophisticated cumbersome teacher model and achieves comparable performance to existing state-of-the-art teacher-free knowledge distillation frameworks when implemented by a low-cost dynamic mannered teacher. Our approach is all-encompassing and can be utilized for any deep neural network training that requires categorization or object recognition. DR-KD enhances the test accuracy on Tiny ImageNet by 2.65\% over prominent baseline models, which is significantly better than any other knowledge distillation approach while requiring no additional training costs.
\end{abstract}



\IEEEpeerreviewmaketitle

\section{Introduction}
Deep learning techniques \cite{lecun2015deep} are used to construct feature hierarchies by linking higher-level features with lower-level features. It is called ``deep" since it comprises multiple steps in the object identification process, all of which are covered during training. 
Convolutional Neural Network (CNN) \cite{krizhevsky2012imagenet} models are a subclass of deep learning models that are used to process tasks such as image classification, image segmentation, object recognition, and natural language processing. Consequently, CNN models have been widely accessible in recent years across a growing number of applications \cite{wang2020deep}. These CNN models demand high efficiency and low latency, with the added requirement to be installed on edge devices.

Edge devices implement edge computing which is a distributed information technology network design that allows mobile computing for locally generated data. Rather than transferring data across the internet to a server for computation, applications compute inputs and operate inside the device itself through the use of a ``small network'' \cite{shi2016edge}.

For example, the rapid expansion of self-driving cars which create a great deal of user data, and patients' personal medical data are regarded to be sensitive pieces of information to be transmitted via the internet owing to the danger of breach. In this environment, edge computing obviates the need for long-distance communication between the client and server, resulting in reduced latency and bandwidth consumption. Subsequently, the speed of computation is as important as the accuracy \cite{xu2019edge}. 

\begin{figure} [H]
    \centering
    \includegraphics[width=0.48\textwidth] {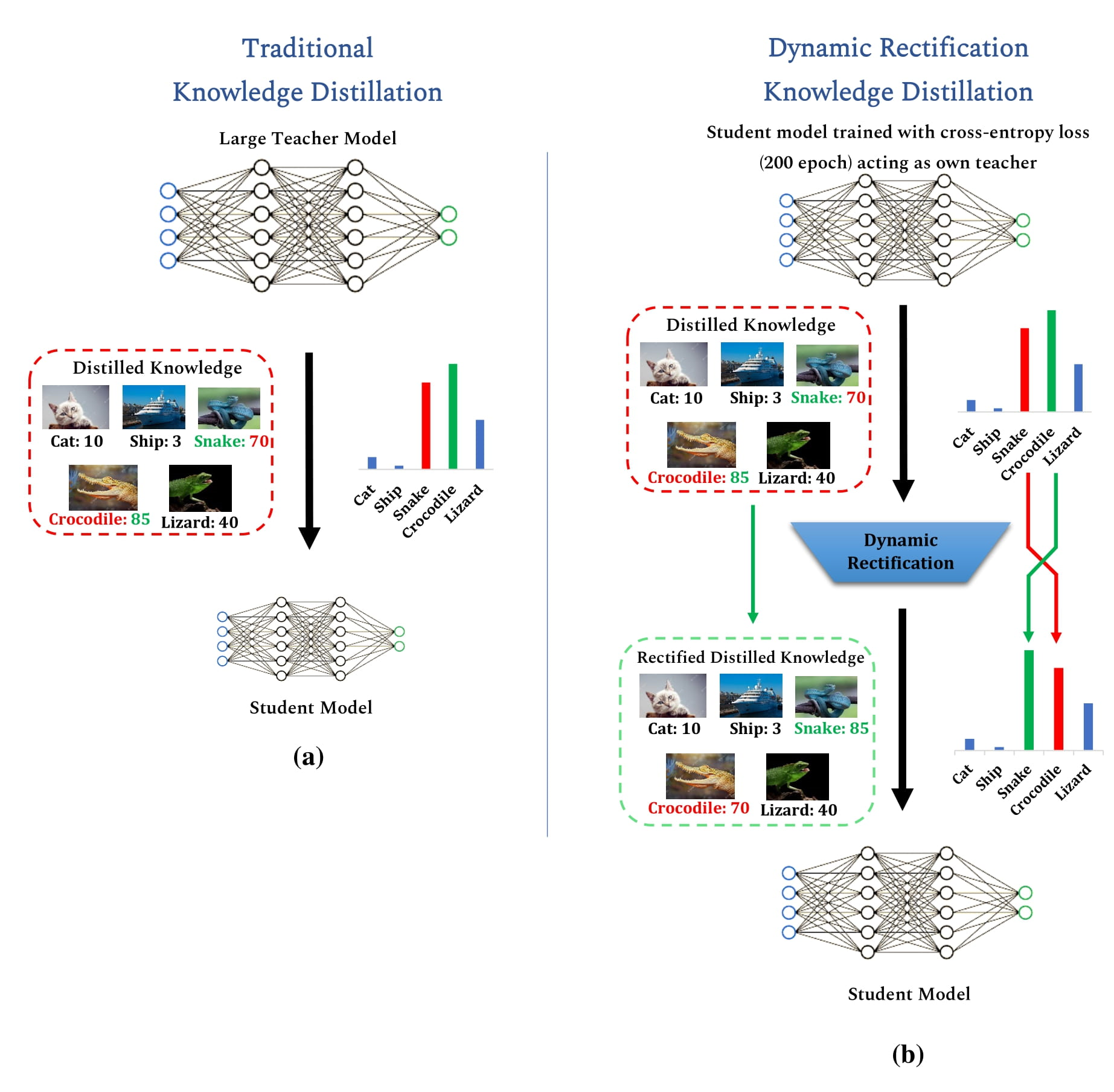}
    \caption{Assume that the input to both of the teacher models in (a) and (b) is the image of a snake. As shown in (a), the teacher model can distill incorrect knowledge to the student model. As seen in (b), the DR-KD technique enables the rectification of incorrect knowledge prior to the knowledge distillation process.}
    \label{fig:picture1}
\end{figure}

A lot of research has concentrated on compressing large models into smaller ones. These studies have resulted in the development of compact models with little to no performance degradation, cited as Deep Model Compression \cite{sau2016deep}.

\begin{figure*} [t]
 \center
  \includegraphics[width=1.0\textwidth]{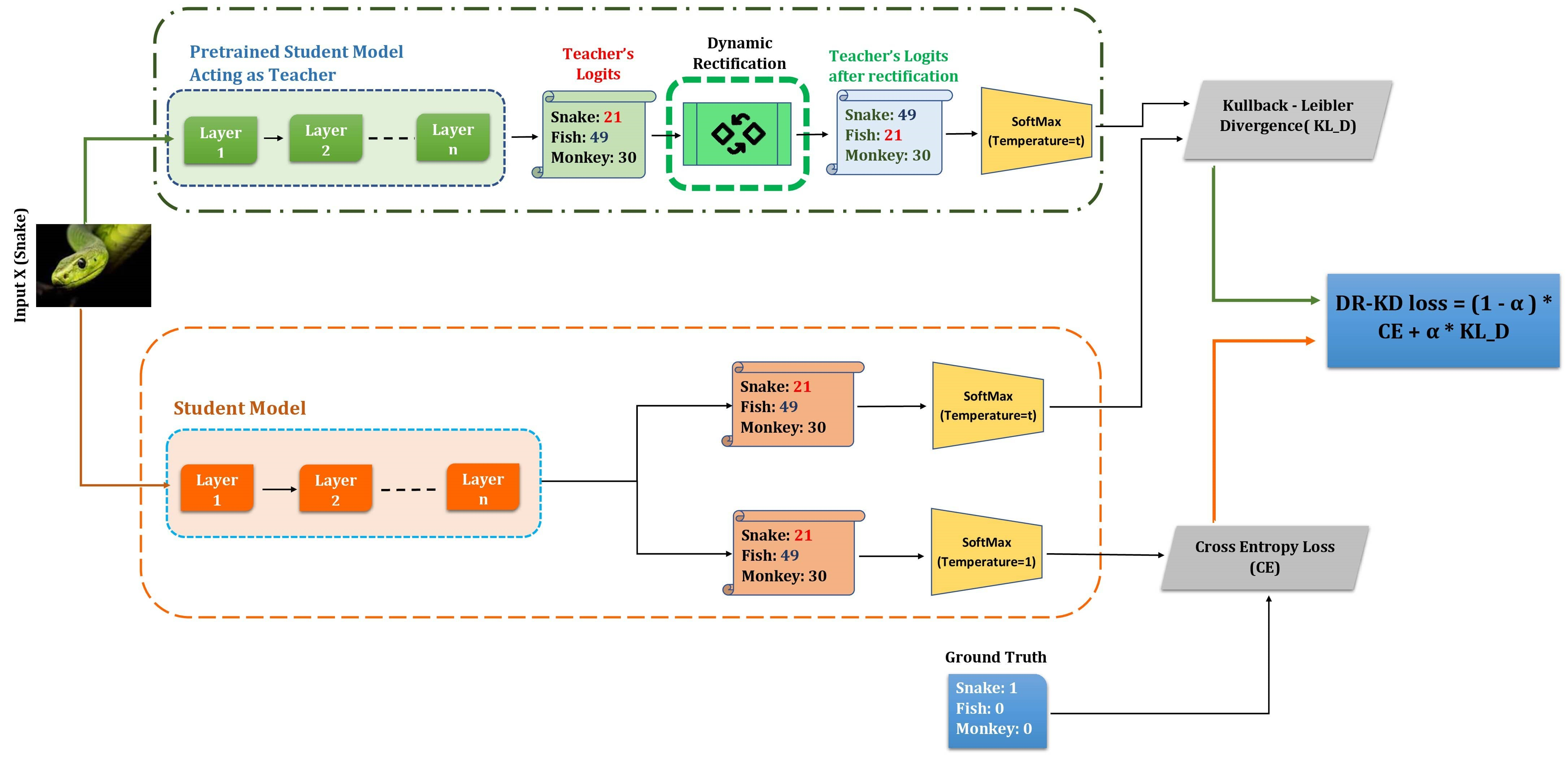}
  \caption{Illustrates the schematic of DR-KD. Both the teacher and student models receive input ``X''. The teacher's incorrectly predicted logit is rectified prior to being passed to the softmax function, but the student logit is directly passed. Kullback-Leibler Divergence (KLD) is determined using the teacher's and student's softmax activation. Concurrently, the logit of the student model is transferred to softmax activation at temperature 1 and the cross-entropy of the student model is determined using ground truth. Finally, the overall loss is computed by combining this with the previously calculated KLD.}
  \label{fig:picture2}
\end{figure*}

Model compression approaches like Knowledge Distillation\cite{hinton2015distilling}, which condense the knowledge of a big complicated model (teacher) and pass it on to train a smaller network (student) to match the output, may be used to meet the accuracy, handle latency and computing requirements at the inference time. The student network is trained to match the prediction of the cumbersome network as well as the distribution of the teacher's network. As a result, the performance of the student model may greatly improve, and may even surpass that of the teacher \cite{deng2020model}. 

However, the primary concern surrounding traditional state-of-the-art knowledge distillation is the inability to develop and train high-performing teacher models. Another limitation is the student model's inability to assimilate all information from the teacher model, which means that high-performing student models with performance equivalent to the teacher model are exceptional.

In this paper, we proposed a knowledge distillation framework which we termed \emph{Dynamic Rectification Knowledge Distillation} \textbf{(DR-KD)} (shown in \textbf{Fig. \ref{fig:picture2}}) to address the drawbacks of traditional distillation techniques and to facilitate the development of an efficient yet highly accurate model. Instead of training a large teacher model and then distilling information from it to the student model, we propose a novel knowledge distillation framework in which training takes place directly on the student model which acts as its own teacher model. However, in this case, we require the teacher to ensure that the knowledge it distills is always accurate using a rectification method. (shown in \textbf{Fig. \ref{fig:picture1}}). This framework requires significantly less training time and resource compared to the traditional student-teacher setup. Our remarkable finding is that the performance of our poorly trained yet rectified teacher model exceeds the performance of a vast highly-trained teacher model by a significant margin. For instance, traditional distillation achieved a test accuracy of 56.70\% on Tiny ImageNet when the teacher was ResNet18 (with 11M parameters)\cite{resnet18cite} and the student was MobileNetV2 (with 3.4M parameters)\cite{mobilenetv2citation}. In contrast, utilizing our technique, MobileNetV2 acts as its own teacher and distills accurate knowledge, achieving a test accuracy of 57.71\% for the same dataset, which is a 2.65\% performance increase over traditional distillation.

In summary, this study makes the following significant contributions:
\begin{itemize}
    \item Our strategy obviates the requirement of ensemble or large model training for teacher models, since a single efficient model may be taught to act as its own teacher. That being the case, substantial training time and computing resources are not required, which is beneficial for real-time data processing and deployment of edge devices.
    \item DR-KD ensures that the teacher never distills incorrect knowledge. This arrangement outperformed the state-of-the-art knowledge distillation techniques available.
    \item DR-KD significantly increases classification task performance without sacrificing response time.
    \item Our proposed framework for knowledge-distillation is platform agnostic. As a result, it may be used in any deep learning pipeline without requiring large modifications.
\end{itemize}

The rest of the paper is structured as follows: Section {\MakeUppercase {\romannumeral 2}} addresses similar studies in this field of research. Section {\MakeUppercase {\romannumeral 3}} delves into the methodological underpinnings of our suggested technique. Section {\MakeUppercase {\romannumeral 4}} is devoted to elucidating the aspects involved in our research methodology. Section {\MakeUppercase {\romannumeral 5}} describes the experimental setup for the tests that we conducted. Section {\MakeUppercase {\romannumeral 6}} discusses our results and performance comparisons. Finally, section {\MakeUppercase {\romannumeral 7}} concludes the paper.

\section{Related Works}

Caruana used the term ``model compression'' to refer to a strategy for training fast, compact models to replicate high-performing but slow and colossal models with minimal performance loss \cite{bucilua2006model}. This idea acquired considerable popularity in subsequent years \cite{ba2013deep}, most notably via the work of Hinton \cite{hinton2015distilling}, whose work is critical to the development of our method. As a result, it is covered in greater depth in Section (\ref{kd-by-hinton}).

Romero et al. introduced FitNet, which included the notion of hint learning, with the goal of minimizing the gap between students' and teachers' feature maps \cite{romero2014fitnets}.

Xu Lan et al. presented a knowledge distillation approach called Self-Referenced Deep Learning (SRDL) \cite{lan2018self} which returns the information collected by the in-training target model to itself in order to regularize the future learning procedure. In comparison to previous knowledge distillation strategies, this one eliminates the need to train a large teacher model and enhances model generalization performance at a low extra computational cost.

Zhang et al. described a novel training approach called self distillation \cite{zhang2019your} that, in comparison to previous distillation methods, removes the need for an extra teacher. Their solution employs an adaptive depth architecture to enable run-time time-accuracy trade-offs. In this study, the authors established their method's superiority over deep supervised networks and prior distillation techniques.

Yang et al. presented a framework named Snapshot Distillation (SD) \cite{yang2019snapshot} as a subset of self-distillation. To facilitate a supervised training process inside the same network, information from the network's earlier epochs (teacher) is transmitted to the network's later epochs (student). This framework achieves the previously unchecked goal of teacher-student (T-S) optimization in a single generation.

Hou et al. demonstrated that their Self-Attention Distillation (SAD) method \cite{hou2019learning} had the potential to significantly boost the visual attention of various layers in diverse networks. They used this method for a lane detection network to enhance its own representation learning without the requirement for extra labels or external supervision and without increasing the base model's inference time.

Phuong and Lampert presented a distillation-based training technique \cite{phuong2019distillation} for reducing inference time via the early exit layer, in which the early exit layer seeks to mimic the output of the later exit layer during training. This proved to be a simple architecture-independent method for multi-exit image classification. It significantly outperforms the state-of-the-art training approach, particularly in contexts with limited data or weak computational resources.

Lee et al. presented a self-distillation technique to be used for data augmentation \cite{lee2019rethinking}. The authors outlined a straightforward yet effective method in which the augmentation's self-knowledge is distilled into the model itself and the model is trained to predict both the label for the original problem and the kind of transformation concurrently.

Zhang and Sabuncu re-examined the technique of multi generational self-distillation and empirically demonstrated that the performance improvement associated with multi generational self-distillation is associated with a growing diversity in teacher predictions \cite{zhang2020self}.

\section{Methodological Background}

In this section we explored, in detail, the studies that are most pertinent to our work. If the reader is familiar with this knowledge, they may proceed directly to Section {\MakeUppercase {\romannumeral 4}}.

\subsection{Knowledge Distillation\label{kd-by-hinton}}

Hinton \cite{hinton2015distilling} proposed knowledge distillation as a method for teaching a weak student to categorize targets defined by an ensemble or large teacher model. The fundamental idea behind this architecture was to distill knowledge from a pre-trained highly accurate teacher model into a smaller weak student model by leveraging the label probabilities produced by the teacher model as ``soft targets''.

To keep things simple, we'll refer to the teacher as model \emph{T} and the student as model \emph{S}. For an input image \emph{x}, $O_t = T(x)$ and $O_s = S(x)$ where $O_t$ refers to teacher model's output score and $O_s$ refers to the student models output score. Both $O_t$ and $O_s$ are then softened using a modified softmax$(\sigma)$ \cite{softmax-cite} with a new hyper-parameter $\tau$ called temperature shown in equation \textbf{(\ref{eqn:softmax})}.

\begin{equation}
\label{eqn:softmax}
\sigma (z_k, \tau) = \frac{e^\frac{Z_k}{\tau}}  {\sum_{m=1}^n {e^\frac{Z_m}{\tau}} }\; \;\;for\:k = 1, 2, ... n
\end{equation}

Following which, the softened $O\textsubscript{t}$ becomes $\mathrm{\emph{O}}_{t}^{\tau}$ and the softened $O\textsubscript{s}$ becomes $\mathrm{\emph{O}}_{s}^{\tau}$. Afterwards, using $\mathrm{\emph{O}}_{t}^{\tau}$ and $\mathrm{\emph{O}}_{s}^{\tau}$, the Kullback–Leibler Divergence (KLD) is calculated. \cite{goldberger2003efficient}. This KLD distance is then combined with the cross-entropy  \cite{cross-entropy} loss of the student model to mimic the teacher by means of minimizing the overall loss shown in equation \textbf{(\ref{eqn:loss_kd})}. 

\begin{equation}
\label{eqn:loss_kd}
    \mathcal{L}_{kd} = (1-\alpha)CE(\sigma(O_s, 1), Y) + \alpha KLD(O_t^{\tau}, O_s^{\tau})
\end{equation}

Here, CE stands for cross-entropy, KLD is for KL divergence, $\alpha$ is a hyper-parameter and $Y$ is the one hot encoded ground truth.

\subsection{Teacher Free knowledge Distillation}
The primary challenge with conventional knowledge distillation is training a cumbersome model to create resource- and time-intensive teacher models. To circumvent this constraint, research in the area of Teacher Free Knowledge Distillation has seen some interest \cite{tfkd}. The concept of this framework was motivated by two further novel experiments conducted by the authors. The first experiment established that a poorly trained teacher model may enhance the performance of a student model, which they termed Defective Knowledge Distillation (De-KD). The second experiment established that a student model can distill knowledge back to a teacher model in order to enhance its performance, which they termed Reversed Knowledge Distillation (Re-KD). By examining the results of these two experiments, the researchers demonstrated that a student model may improve its performance without the assistance of a teacher model completely, which they termed Teacher Free Knowledge Distillation (Tf-KD) \cite{tfkd}.

The first Tf-KD approach is referred to as ``Self-Training Knowledge Distillation'' or Tf-KD\textsubscript{self}. As the authors explain in their study, a student may teach a teacher, and a poorly trained teacher can benefit a student as well. The authors devised this strategy in response to the absence of a large teacher model. They trained a model with the configuration of student $S$  over  a  certain  number  of  epochs using cross-entropy loss to build a pre-trained self-teacher model that was then utilized as the weak teacher $S_t$ in order to distill knowledge to the student $S$ using soft target transfer. In their experiment, the authors employed Hinton's\cite{hinton2015distilling} knowledge distillation setup, including the loss function,   described in section (\ref{kd-by-hinton}), but they substituted the cumbersome teacher $T$ with the weak teacher $S_t$. This means the output scores $O_t=S_t(x)$ instead of coming from $T$. Meaning now $O_t^\tau=\sigma(S_t(x),\tau)$ , that is the KL divergence term learns from the weak teacher.


\section{Dynamic Rectification Knowledge Distillation}

When a small model is trained to generate a teacher model, it may misclassify samples if it is under-trained. If this configuration is maintained and the teacher model makes an inaccurate prediction, then the student model will be impacted by this erroneous data through the KL divergence component of the loss function mentioned in equation (\textbf{\ref{eqn:loss_kd}}). If required, for instance, when the teacher makes an error, we suggest that information should be rectified prior to being provided to the student. This rectification process can take a number of different forms. In this paper, we propose a simple yet efficient rectification technique as proof of concept.

We maintained a comparable process to Hinton's knowledge distillation framework where there is a teacher model whose knowledge will be distilled into the student. Also like Tf-KD\textsubscript{self}, we trained the student model $S$ conventionally to generate a pre-trained teacher model $S_t$. However, we have sought to improve performance of the teacher by correcting instances of incorrect predictions using a swapping method which we termed \emph{Dynamic Rectification (DR)} shown in equation \textbf{(\ref{eqn:dr-kd-swap})}. Instead of using the teacher logits directly, we cross-checked the logits against the labels to determine whether the highest logit from the teacher’s prediction mapped to the label’s ground truth. If not, we used our $DR$ function to swap the logits in order to distill the correct information as knowledge.

The concept behind the rectification process is as follows: if the highest logit in the $S_t$ model indicates an inaccurate class, we will swap the accurate class's logit with the maximum logit of that  inaccurate class. This will result in the accurate class indicating the highest logit.
Using this strategy, the $S_t$ model will always be the better teacher and capable of performing effective knowledge distillation, while enabling the $S$  model to learn more effectively.

The output score of $S$ is $P_s$ and the output score of $S_t$ is $P_t$. $P_s$  is directly softened using the modified softmax (equation \textbf{(\ref{eqn:softmax})}) with temperature $\tau$ to generate $P_s^\tau$ = $\sigma(P_s, \tau)$. However $P_t$ is first rectified through the rectification function in equation \textbf{(\ref{eqn:dr-kd-swap})} to generate $P_{t\_rect}=DR(P_t,Y)$. Here $Y$ is the one hot encoded ground truth, $j=argmax(Y)$ which is the index of the maximum value of $Y$.Similarly $l=argmax(P_t)$ which is the index of the maximum value of $P_t$. Later on the $P_{t\_rect}$ is passed to the modified softmax with temperature $\tau$ to generate $P_{t\_rect}^\tau$.

\noindent
\begin{equation}
\label{eqn:dr-kd-swap}
DR(P_t,Y)=\begin{cases}
          P_t=P_t \quad &\text{if} \,  \emph{j} = \emph{l} \\[5mm]
          \hbar=P_t(j) \\ P_t(j)=P_t(l) \quad &\text{if} \,\emph{j} \neq \emph{l} \\ P_t(l)=\hbar
     \end{cases}
\end{equation}

 
Then $P_s^\tau$ and $P_{t\_rect}^\tau$ are used to calculate the KL divergence. Later on this KL divergence is combined with the cross entropy loss to calculate the total weighted loss as shown in equation \textbf{(\ref{eqn:dr-kd-loss})} where as before CE is cross entropy and $\alpha$ is hyperparameter to weight the different component of the loss
 
\vspace*{-\abovedisplayskip}
\begin{equation}
\label{eqn:dr-kd-loss}
    \mathcal{L}_{DR-KD} = (1-\alpha)CE(\sigma(P_s, 1), Y) + \alpha KLD(P_{t\_rect}^\tau, P_s^{\tau})
\end{equation}



\section{Experimental Setup}
We conducted experiments to evaluate the performance of DR-KD on three major publicly available image classification datasets namely CIFAR-10 \cite{cifar10}, CIFAR-100 \cite{cifar10} and Tiny ImageNet \cite{tiny-imagenet} and evaluated its performance. For our trials on the three datasets, we picked a total of six models named MobileNetV2, ShuffleNetV2 \cite{ma2018shufflenet}, ResNet18 \cite{resnet18cite}, ResNet50 \cite{resnet50}, GoogLeNet \cite{googlenetcite} and DenseNet121\cite{densenet121cite}. For fair comparability, we employed the same training environment as in Tf-KD for all datasets. \textbf{Table \ref{tab:hyperparameter}} shows the setup of hyperparameters for our experiments.

All experiments were done on Google Colab. The source code is made publicly available and can be accessed \href{https://github.com/Amik-TJ/dynamic_rectification_knowledge_distillation}{\color{blue}here}\footnote{\url{https://github.com/Amik-TJ/dynamic_rectification_knowledge_distillation}}.

\begin{table}[H]
\centering
\caption{Hyperparameters: Temperature $\tau$ and $\alpha$ for DR-KD experiments}
\label{tab:hyperparameter}
\begin{tabular}{ccc}
\hline
\textbf{Dataset} &
  \textbf{Model} &
  \textbf{DR-KD} \\ \hline
\begin{tabular}[c]{@{}c@{}}CIFAR-10 \& \\ CIFAR-100\end{tabular} &
  \begin{tabular}[c]{@{}c@{}}MobileNetV2\\ ShuffleNetV2\\ ResNet18\\ GoogLeNet\end{tabular} &
  \begin{tabular}[c]{@{}c@{}}$\tau$=20, $\alpha$=0.95\\ $\tau$=20, $\alpha$=0.95\\ $\tau$=6, $\alpha$=0.95\\ $\tau$=20, $\alpha$=0.40\end{tabular} \\ \hline
Tiny-ImageNet &
  \begin{tabular}[c]{@{}c@{}}MobileNetV2\\ ShuffleNetV2\\ ResNet50\\ DenseNet121\end{tabular} &
  \begin{tabular}[c]{@{}c@{}}$\tau$=20, $\alpha$=0.10\\ $\tau$=20, $\alpha$=0.10\\ $\tau$=20, $\alpha$=0.10\\ $\tau$=20, $\alpha$=0.15
  \end{tabular}
\\ \hline
\end{tabular}

\end{table}

\section{Results}

We compared DR-KD's test accuracy with the following available knowledge distillation frameworks: Tf-Kd\textsubscript{self}, Teacher free knowledge distillation via regularization (Tf-KD\textsubscript{reg}), Label Smoothing Regularization (LSR) \cite{label-smoothing} \cite{tfkd}, Normal KD \cite{hinton2015distilling}. The term “baseline” refers to traditional CNN training with just cross-entropy loss.

\begin{table*}[t]
 \centering
 \caption{Test accuracy on CIFAR-10. DR-KD improves performance of all four models by a significant margin. The values shown in bold are the best scores.}
 \label{tab:cifar10-table}
\begin{tabular}{lcccc}
\hline
\multicolumn{1}{c}{

\textbf{Model}} &
  \textbf{Baseline} &
  \textbf{DR-KD} &
  \textbf{\begin{tabular}[c]{@{}c@{}}Increment (DR-KD - Baseline)\end{tabular}} &
  \textbf{Increment in \%} \\ \hline
MobileNetV2  & 90.72 & \textbf{91.77} & +  1.05 & 1.16\% \\
ShuffleNetV2 & 91.55 & \textbf{92.74} & + 1.19  & 1.30\% \\
Resnet18     & 94.96 & \textbf{95.30} & + 0.34  & 0.36\% \\
GoogLeNet    & 95.15 & \textbf{95.63} & + 0.48  & 0.50\% \\ \hline
\end{tabular}

\end{table*}

\begin{table*}[t]
\centering
\caption{Test accuracy in \% on CIFAR-100. All experiments were repeated four times.  The values shown in bold are the best scores.}
\label{tab:cifar100-table}
\begin{tabular}{lcccccl}
\hline
\multicolumn{1}{c}{\textbf{Model}} &
  \textbf{Baseline} &
  \textbf{DR-KD} &
  \textbf{Tf-KD\textsubscript{self}} &
  \textbf{Tf-KD\textsubscript{reg}} &
  \textbf{+LSR} &
  \multicolumn{1}{c}{\textbf{Normal KD {[}Teacher{]}}} \\ \hline
MobileNetV2  & 68.38 & \textbf{71.66 $\pm$ 0.21} & 70.96 $\pm$ 0.24 & 70.88 & 69.32 & 71.05 {[}ResNet18{]}  \\
ShuffleNetV2 & 70.34 & \textbf{72.59 $\pm$ 0.25} & 72.23 $\pm$ 0.16 & 72.09 & 70.83 & 72.05 {[}ResNet18{]}  \\
Resnet18     & 75.87 & \textbf{77.47 $\pm$ 0.13} & 77.10 $\pm$ 0.20 & 77.36 & 77.26 & 77.19 {[}ResNet50{]}  \\
GoogLeNet    & 78.72 & \textbf{80.25 $\pm$ 0.39} & 80.17 $\pm$ 0.15 & 79.22 & 79.07 & 78.84 {[}ResNeXt29{]} \\ \hline
\end{tabular}
\end{table*}

\begin{table*}[t]
\centering
\caption{Performance improvement in \% on CIFAR-100. For each KD framework, the accuracy improvement was determined by subtracting the baseline performance from the framework’s performance score. The values shown in bold are the best scores.}
\label{tab:cifar100-comparison-table}
\begin{tabular}{lccccc}
\hline
\multicolumn{1}{c}{\textbf{Model}} & 
\textbf{DR-KD} & 
\textbf{Tf-KD\textsubscript{self}} &
  \textbf{Tf-KD\textsubscript{reg}} &
\textbf{LSR} & 
\textbf{Normal KD} \\ \hline
MobileNetV2  & \textbf{+3.28} & +2.58 & +2.50 & +0.94 & +2.67 {[}ResNet18{]}  \\
ShuffleNetV2 & \textbf{+2.25} & +1.89 & +1.74 & +0.49 & +1.71 {[}ResNet18{]}  \\
Resnet18     & \textbf{+1.60} & +1.23 & +1.49 & +1.39 & +1.19 {[}ResNet50{]}  \\
GoogLeNet    & \textbf{+1.53} & +1.45 & +0.50 & +0.92 & +1.39 {[}ResNeXt29{]} \\ \hline
\end{tabular}
\end{table*}

\begin{figure} [H]
    \label{fi}
    \centering
    \includegraphics[width=0.47\textwidth] {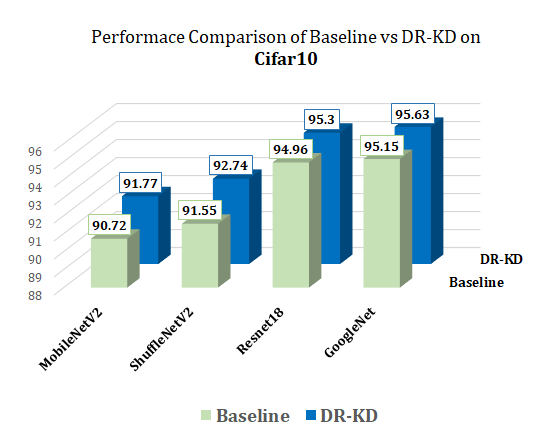}
    \caption{CIFAR-10 test accuracy in \%}
    \label{fig:picture3}
\end{figure}

\subsection{Experiment Results on CIFAR-10}

CIFAR-10 contains 60000 units of RGB images of which, 50000 are allocated for training and the rest are allocated for testing. There are a total of 10 classes and each class contains 6000 images.

We built baseline models for CIFAR-10 by training the MobileNetV2, ShuffleNet, ResNet18, and GoogLeNet for 200 epochs so that we could utilize them as poorly trained teacher models. The batch size was 256 for MobileNetV2 and ShuffleNet and 128 for ResNet18 and GoogLeNet.

\textbf{Table \ref{tab:cifar10-table}} shows that  DR-KD  surpassed  all  baseline  scores by  a  fair  margin  and  outperformed the other Knowledge Distillation  techniques without  any  training  required  by  the cumbersome model. For instance, performance of ShuffleNet increased by a noteworthy 1.3\%. GoogLeNet, an ensemble model composed of 22 deep layers, increased its test accuracy from 95.15\% to 95.63\% with DR-KD.

\textbf{Fig. \ref{fig:picture3}} illustrates our models' test accuracy. As it can be observed,  DR-KD outperforms the baselines consistently. For instance, as a mobile-compatible model with 3.4M parameters, MobileNetV2 improves its baseline score from 90.72 to 91.77, which is a 1.16\% gain.

\subsection{Experiment Results on CIFAR-100}

CIFAR-100 is comparable in size to CIFAR-10, with the exception that it has 100 classes. These classes each have 500 training images and 100 testing images. 

\textbf{Table \ref{tab:cifar100-comparison-table}} shows that the 11M parameter containing Resnet18 outperforms baseline by \textbf{1.60}\% which is substantially greater than the other four Knowledge Distillation frameworks. Based on these results, it is evident that our technique works significantly better than the existing state-of-the-art knowledge distillation frameworks.

\begin{figure} [H]
    \centering
    \includegraphics[width=0.48\textwidth] {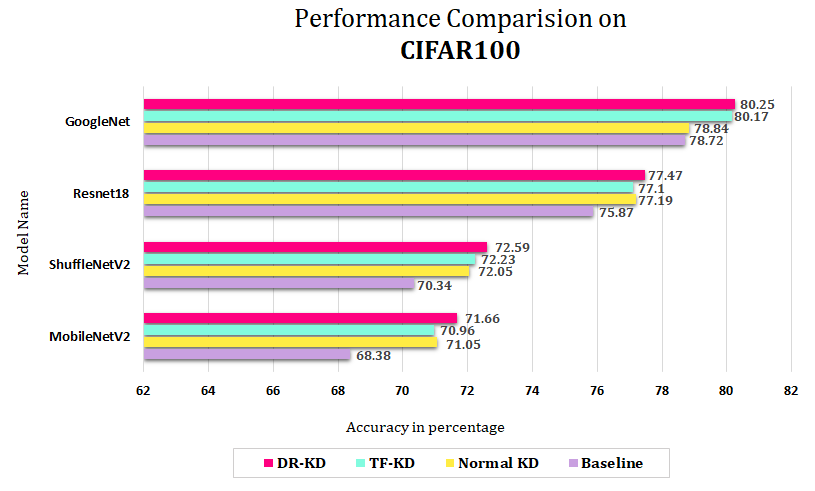}
    \caption{Performance Comparison on CIFAR-100}
    \label{fig:picture4}
\end{figure}

\begin{figure} [H]
    \centering
    \includegraphics[width=0.48\textwidth] {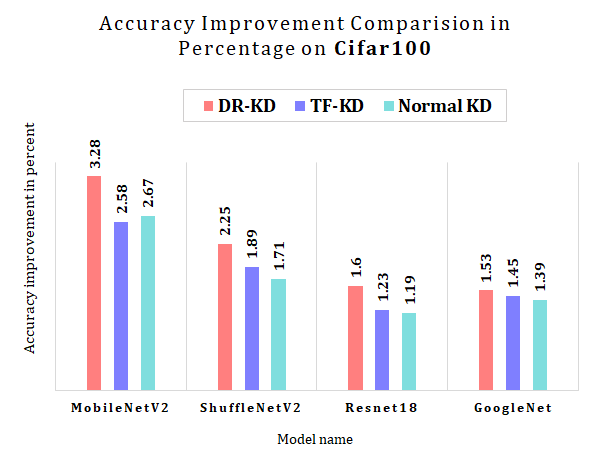}
    \caption{Accuracy Improvement Comparison in \% on CIFAR-100}
     \label{fig:picture5}
\end{figure}

DR-KD's performance markedly surpassed all of the benchmarks, including baseline, Tf-KD\textsubscript{self}, Tf-KD\textsubscript{reg}, LSR and Normal KD, as shown in \textbf{Fig. \ref{fig:picture4}}. When applied to poorly trained teacher models, implementation of DR-KD resulted in a 3.28\% performance improvement for MobileNetV2 shown in \textbf{Fig. \ref{fig:picture5}}.

\begin{table*}[t]
\centering
\caption{TINY IMAGENET results. Increased test accuracy above baseline is shown by the number in parenthesis.}
\label{tab:tiny-imagenet-table}
\begin{tabular}{ccccccl}
\hline
\textbf{Model} &
  \textbf{Baseline} &
  \textbf{DR-KD} &
  \textbf{Tf-KD\textsubscript{self}} &
  \textbf{Tf-KD\textsubscript{reg}} &
  \textbf{LSR} &
  \multicolumn{1}{c}{\textbf{Normal KD{[}Teacher{]}}} \\ \hline
MobileNetV2  & 55.06 & \textbf{57.71 (+2.65)}  & 56.77 (+1.71) & 56.47 (+1.41) & 56.24 (+1.18) & 56.70 (+1.64) {[}ResNet18{]}    \\ 
ShuffleNetV2 & 60.51 & \textbf{61.63 (+1.12)}  & 61.36 (+0.85) & 60.93 (+0.42) & 60.66 (+0.11) & 61.19 (+0.68) {[}ResNet18{]}    \\
ResNet50     & 67.47 & \textbf{68.26 (+0.79)}  & 68.18 (+0.71) & 67.92 (+0.45) & 67.63 (+0.16) & 68.23 (+0.76) {[}DenseNet121{]} \\
DenseNet121  & 68.15 & \textbf{68.35 (+0.20)} & 68.29 (+0.14) & 68.37 (+0.18) & 68.19 (+0.04) & 68.31 (+0.16) {[}ResNeXt29{]} \\ \hline 
\end{tabular}
\end{table*}

\subsection{Experiment results on Tiny ImageNet}

The final dataset that we used for our experiments was Tiny ImageNet, which is a 200-class subset of ImageNet. Out of a total of 110000 images, 100000 are in the training set and the remainder are in the test set.

On Tiny ImageNet, we trained MobileNetV2, ShuffleNetV2, ResNet50 and DenseNet121 with cross-entropy loss to generate the baseline models. The batch size is 64 for larger models namely ResNet50 and DenseNet121 and the batch size is 128 for the other two small models. All of them are trained over a period of 200 epochs. 

\textbf{Table \ref{tab:tiny-imagenet-table}} shows the results of Baseline, DR-KD, Tf-KD\textsubscript{self}, Tf-KD\textsubscript{reg}, LSR, and Normal KD on Tiny-ImageNet. When compared to other Knowledge Distillation frameworks, DR-KD consistently and significantly improves the performance of all models. Among all the experiments, DR-KD's notable attainment is for MobileNetV2 on Tiny ImageNet. The baseline accuracy for MobileNetV2 is 55.06\% whereas DR-KD’s test accuracy is 57.71\%. This indicates that DR-KD achieves an increment of 2.65\% compared to 1.71\% for Tf-KD\textsubscript{self}, 1.41\% for Tf-KD\textsubscript{reg}, 1.18\% for LSR and 1,64\% for Normal KD.  In terms of distilling knowledge to ShuffleNetV2, our rectified self-taught ShuffleNetv2 teacher model (with approximately 2M parameters) substantially outperforms the ResNet18 teacher model (with approximately 11M parameters).

We were unable to conduct any experiments on the ImageNet \cite{imagenet} dataset due to lack of computational resources.

\section{Conclusion}

Traditional knowledge distillation necessitates the training of a very large teacher model which may prove to be costly both in terms of processing time and money. There have been proposed approaches to self-distillation that totally eliminate the need for a teacher model, as well as approaches that train a poorly trained teacher which provide acceptable results. In this paper, we propose a novel method in which we take a low-resource weak teacher model and then process its outputs  through a rectification mechanism to train the student model. To measure the efficacy of our method we evaluate the performance of six models on three datasets. We demonstrate that this very simple operation yields results which are significantly better than previously published studies. Most notably on Tiny ImageNet our DR-KD technique improves test accuracy by 2.65\% and on CIFAR100 it exceeds the baseline by 3.28\%. In addition to advancing technology, our work benefits Machine Learning practitioners who lack access to high-performance computers. Larger datasets, like ImageNet, remain untrainable on Google Colab. Conventionally, in order to train a small model, we must train a large model first which requires expensive resources. This technique enables us to achieve significant performance with only a small model makes machine learning economically viable for the masses.






\bibliographystyle{IEEEtran}
\bibliography{myBib.bib}
%


\end{document}